\documentclass[letterpaper, 11 pt, conference]{ieeeconf}  

\IEEEoverridecommandlockouts                              

\usepackage{cite}
\usepackage{booktabs}
\usepackage{graphicx}
\usepackage{subcaption}
\usepackage{amsmath}
\usepackage{multicol, blindtext}
\usepackage{authblk}
\usepackage{hyperref}
\usepackage{caption}
\usepackage{multirow}
\usepackage{siunitx}
\usepackage{etoolbox}
\usepackage{amsfonts}
\usepackage{multirow}
\usepackage{url}
\usepackage[labelsep=period]{caption}
\usepackage[style=ieee]{}
\title{\LARGE\bf Multimodal Aggregation Approach for Memory Vision-Voice Indoor Navigation with Meta-Learning}

\author{$^{\dagger}$Liqi Yan$^{1,3}$, $^{\dagger}$Dongfang Liu$^{2}$, Yaoxian Song$^{1,3}$, $^{*}$Changbin Yu$^{3}$ 
\thanks{This work is supported in part by the National Science Foundation China (NSFC)
61761136005, and ARC through DP190100887 and DP160104500.} 
\thanks{$^{\dagger}$These authors contributed equally to this work.}
\thanks{$^{*}$Corresponding email: {\tt\small yu\_lab@westlake.edu.cn}}
\thanks{$^{1}$L. Yan and Y. Song are enrolled at Fudan University, China
        }
\thanks{$^{2}$D. Liu is with Department of Computer Graphics Technology,
        Purdue University, USA
        }
\thanks{$^{3}$L. Yan, Y. Song and C. Yu are with the School of Engineering,
        Westlake University, China
        }

}

\begin{document}

\maketitle
\thispagestyle{empty}
\pagestyle{empty}

\begin{abstract}
Vision and voice are two vital keys for agents' interaction and learning. In this paper, we present a novel indoor navigation model called Memory Vision-Voice Indoor Navigation (MVV-IN), which receives voice commands and analyzes multimodal information of visual observation in order to enhance robots' environment understanding. We make use of single RGB images taken by a first-view monocular camera. We also apply a self-attention mechanism to keep the agent focusing on key areas. Memory is important for the agent to avoid repeating certain tasks unnecessarily and in order for it to adapt adequately to new scenes, therefore, we make use of meta-learning. We have experimented with various functional features extracted from visual observation. Comparative experiments prove that our methods outperform state-of-the-art baselines.
\end{abstract}

\begin{figure}[htbp]
	\centerline{\includegraphics[width=3.0in]{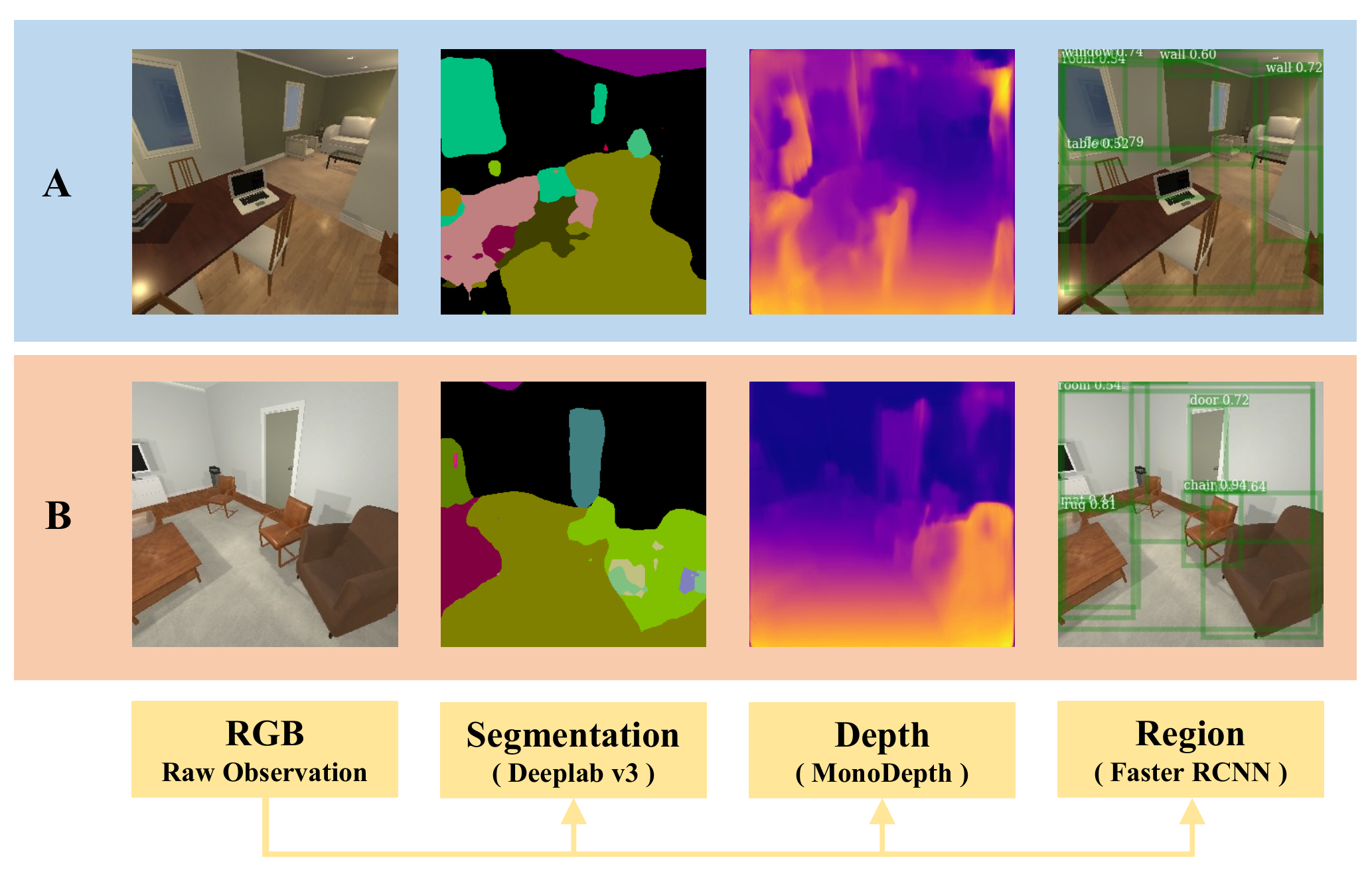}}
	\caption{Two demos of semantic segmentation, depth estimation and object detection from visual observation. Semantic segmentation can identify walkable areas, walls and obstacles, while object detection recognizes targets and depth estimation predicts the distance of all of the above.}
	\label{SegDepthFig}
\end{figure}

\section{Introduction} 
\indent Indoor navigation is an indispensable ability of mobile robotic systems to operate on daily tasks. In order to fulfill various missions, it is imperative for the indoor mobile robot to be able to realize highly efficient search, plus location and approach tasks in restricted circumstances \cite{anderson2018evaluation} \cite{liu2019virtual}. The above capabilities are conducive to a lot of practical applications, enabling higher living standards and enhancing human competence. For instance, such capabilities would enable personal assistant and hazard-extinguishing robots to operate more efficiently under indoor circumstances.\\
\indent Aiming to make a leap forward in the field of navigation models, many in-depth reinforcement learning (RL) methods have been put forward 
\cite{wortsman2019learning}\cite{zhu2017target}\cite{ li2019walking}. Some people describe these as the Partially Observable Markov Decision Process (POMDP) and end-to-end policy learning approaches, which are used to connect observation and movements \cite{mousavian2019visual}\cite{wang2018look} \cite{wang2019reinforced}\cite{savva2017minos} \cite{wu2018building}. Yet, for all that, RL algorithms frequently suffer from data hunger 
for training. A large array of 3D synthetic environments have been created to resolve this issue but the trained models sometimes are unworkable in practical application \cite{yu2018guided} due to the fact that they struggle to adapt to varied circumstances \cite{gupta2017cognitive} \cite{mirowski2016learning}.\\
\indent There is an inherent chaos in daily life, which deserves our attention. For new daily missions, people are able to find solutions within a relatively short time by applying meta-skills acquired from multiple past activities and experiences. This phenomenon provides an inspiring analogy with the present RL-oriented navigation approaches. Human beings are born with the innate ability to transfer prior knowledge, to learn new skills from experience and to apply them to a variety of missions freely. We aim for machines and robots to be equipped with similar capacities.\\
\indent Bearing this in mind, we focus on meta-learning \cite{nichol2018first} \cite{finn2017model} and transfer learning \cite{torrey2010transfer} \cite{weiss2016survey}. The former can realize efficient learning with a reduced amounts of data, while the latter could facilitate the learning of new tasks via the application of knowledge already acquired from relevant familiar activities. In contrast, those other representative meta-learning approaches \cite{nichol2018first} \cite{finn2017model} require large amounts of manually designed missions, which is unfeasible for embodied agents.\\
\indent In this paper, we introduce a novel RL unsupervised method, capable of generating various missions. As mentioned, we call the proposed method, the Memory Vision-Voice Indoor Navigation (MVV-IN). In our design, the agent could acquire transferable meta-skills proficiently so, as to cope with new situations by applying meta-skills to navigate them. As Fig.~\ref{SegDepthFig} shows, our model uses multi-modal aggregation, which integrates voice target information into the network and also integrates spatial and semantic information obtained by monocular vision. In addition, we apply advanced attention mechanisms to the agent's perception of the object region, so that the same can focus on important objects. The model includes memory able to remember short and even long-term states, so, as to avoid the loss of important relevant information. We perform our experiments in various settings and experimental results show that the method significantly outperforms the baseline in all of the metrics and in inference speed.  \\
\indent In summary, contributions of our work include: (a) a proposal of a new navigation task combining vision and voice, (b) the feeding and aggregation of multimodal visual information into the navigation network, (c) the proposal of region self-attention in visual environment understanding and navigation, (d) robust experimental proof that multimodal visual information improves significantly the navigation performance.
\section{Related Work}

\indent The advancement of computer vision has opened the door to considerable effort directed to use deep-learning approaches for indoor navigation. \cite{zhu2017target} has created a type of indoor navigation, in which the mobile robot is provided with the image of its object; \cite{mirowski2016learning} has solved difficulties in indoor navigation via the training of a dual mapper and planner model; \cite{gupta2017cognitive} accelerates RL training of indoor navigation under the auspices of loop closure detection; \cite{savinov2018semi} added a topological map to his study of indoor navigation missions; \cite{kahn2018self} devised a deep RL model that could realize autonomous supervision for navigation; \cite{toshev2018visual} anticipated navigation strategies by means of target inspectors and semantic segmentation models;and \cite{wu2018learning}\cite{yang2018visual} have integrated the semantic information to improve the generalization for unseen settings. All the above perspectives of research have provided reference to our studies.\\
\indent Our approach differs from others in the specific updated parameters used during navigation and the ways to achieve it. Applications as autonomous drive \cite{liu2020video}, map-oriented city navigation \cite{cummins2007probabilistic} and gameplay \cite{wu2017scalable} have been probed into learning-centered navigation. In addition, language-training-based navigation has been studied \cite{konolige2008outdoor}\cite{yu2018interactive} \cite{zhang2016colorful}. In our paper, we aim at applying meta-learning to navigate new scenes more successfully.

\begin{figure*}[htbp]
	\centerline{\includegraphics[width=6.8in]{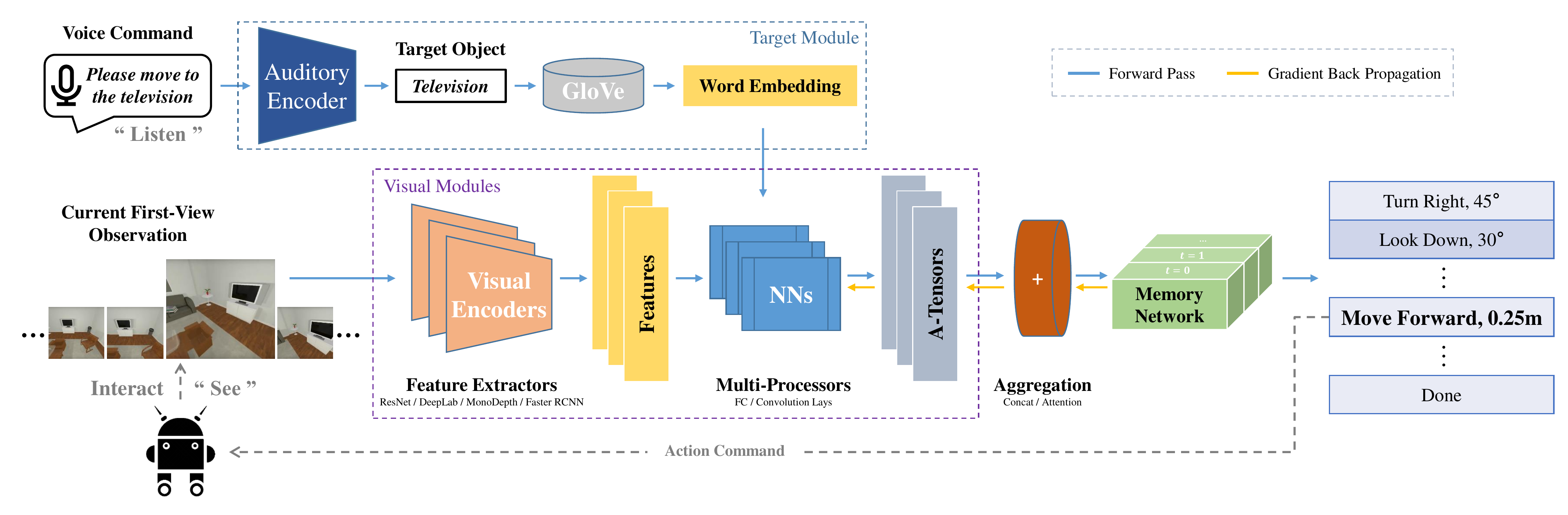}}
	\caption{The pipeline of our proposed MVV-IN system. The current visual observation is input to different encoders to extract different features, like RGB feature and depth feature. After multi-processors, aggregatable tensors (A-Tensor) converted from those features and word embedding, are aggregated and inputted to the memory network. According to its internal state, the memory cell outputs an action policy and its value, like "Move Forward by 0.25m", which will be executed by the agent.  }
	\label{OverallFig}
\end{figure*}

\section{Our Approach}
\subsection{Task Definition}
\indent \textbf{Vision-Voice Navigation task.} Given a command by voice $ V $, e.g. ``\textit{Please move to the television}", our goal is to navigate in a room. The agent should start from a random location $ l_0 $ and use only RGB pictures taken by monocular camera from the first perspective. Inspired from Seq2Seq (Sequence-to-Sequence) \cite{Sutskever2014Sequence} model, we formulate the vision-voice task $ t_{mvv-in} \in T_{mvv-in} $ as a tuple $ t_{mvv-in} = (V,l_0,r_0) $, where $ r_0 $ is the random starting orientation. Generally, the MVV-IN system receives the voice signal $ V $ as well as a sequence of images $ \mathcal{I}=(I_0,...,I_q) $ when the agent interacts with the environment, and carries out a sequence of actions as policy $ \mathcal{P}=(a_0, ..., a_q) $ after $ q+1 $ steps, where $ a \in A $ and $ A $ is a set of actions the agent can perform. The system estimates the probability:
\begin{equation}
    P(\mathcal{P} | V, l_0, r_0) = \prod_{i=0}^q P(a_i | V, a_{<i}, I_{<i})
\end{equation}{}where  $ I_{<i} $ and $ a_{<i} $ represents all the visual observations and issued actions before the $ i^{th} $ step. The issued proposal $ \mathcal{P} $ with the highest estimate should be able to reach the target object with the minimum steps.

\indent \textbf{Meta-learning task.} Meta-learning is to learn how to train a model. We need to train and test the model for visual-voice navigation separately under various scenes (rooms). A complete training and testing process is called a meta learning task. Formally, suppose there are $ q $ scenes in total, we define a set of scenes $ S=\{s_0,...s_q\} $, so the meta-learning task $ \tau_{ml} \in \mathcal{T}_{ml} $ can be denoted by the tuple $ \tau_{ml}=(s,t_{mvv-in}) $, where $ s \in S $ and $ t_{mvv-in} $ is the MVV-IN task we defined earlier. We divide $ S $ into two disjoint sets $ S_{tr} $ and $ S_{te} $, for the training tasks $ \mathcal{T}_{tr} $ and $ \mathcal{T}_{te} $ respectively. Our task is to let the model find the initialization parameter values that are good for training and testing in all different scenes. 

\subsection{Model Design}
\indent Our model is modified from the baseline SAVN \cite{wortsman2019learning} architecture, which can be described with encoder-decoder \cite{Cho2014On} architecture, as shown in Fig.~\ref{OverallFig}. Encoder can be divided into two types: auditory encoder and visual encoder. Both of their output needs to be converted into A-tensor through the processors. Compared with the original network structure, the main contribution of our paper is the design of multi-processor, which can realize multi-modal fusion and attention mechanism. 

\indent The auditory encoder is to recognize the object target from the speech voice. It consists of two parts, one is language recognition module, the other is target extraction module. The first is responsible for translating the language into the text, while the second is responsible for extracting the target $ o \in O $ from the text according to the given rules. 

\indent The visual encoder is to mine the deep information of visual observation and extract specific feature from it. We use several different visual analysis decoders to extract different visual features. These features include general RGB feature $ F^{(RGB)} $, semantic segmentation feature $ F^{(s)} $, depth feature $ F^{(d)} $, and object region feature $ F^{(o)} $. At the meantime, object region-box proposals $ B^{(o)} $ are also predicted and aggregated in our model. 

\indent As Fig.~\ref{OverallFig} shows, our visual module contain the visual encoder as well as the multi-processor, which is to convert the feature of image into an aggregable tensor (A-Tensor). Those A-Tensors is to be mix together and condensed into an embedding using multimodal aggregation. Formally, the embedding obtained by the multimodal aggregation method can be denoted as: 
\begin{equation}
    e_i=\gamma (F_i^{(RGB)}, F_i^{(s)}, F_i^{(d)}, F_i^{(o)}, B^{(o)})
\end{equation}{}
\indent The decoder takes the responsibility of memory. So we utilize memory network as decoder, whose mission is to choose the action with the maximal probability at step $ i $ under the condition $ (h_i,e_i,a_{i-1}) $ and the network parameters, where $ h_i $ is the stored hidden state. 

\subsection{Multimodal Aggregation} 
\begin{figure}[htbp]
	\centerline{\includegraphics[width=3.2in]{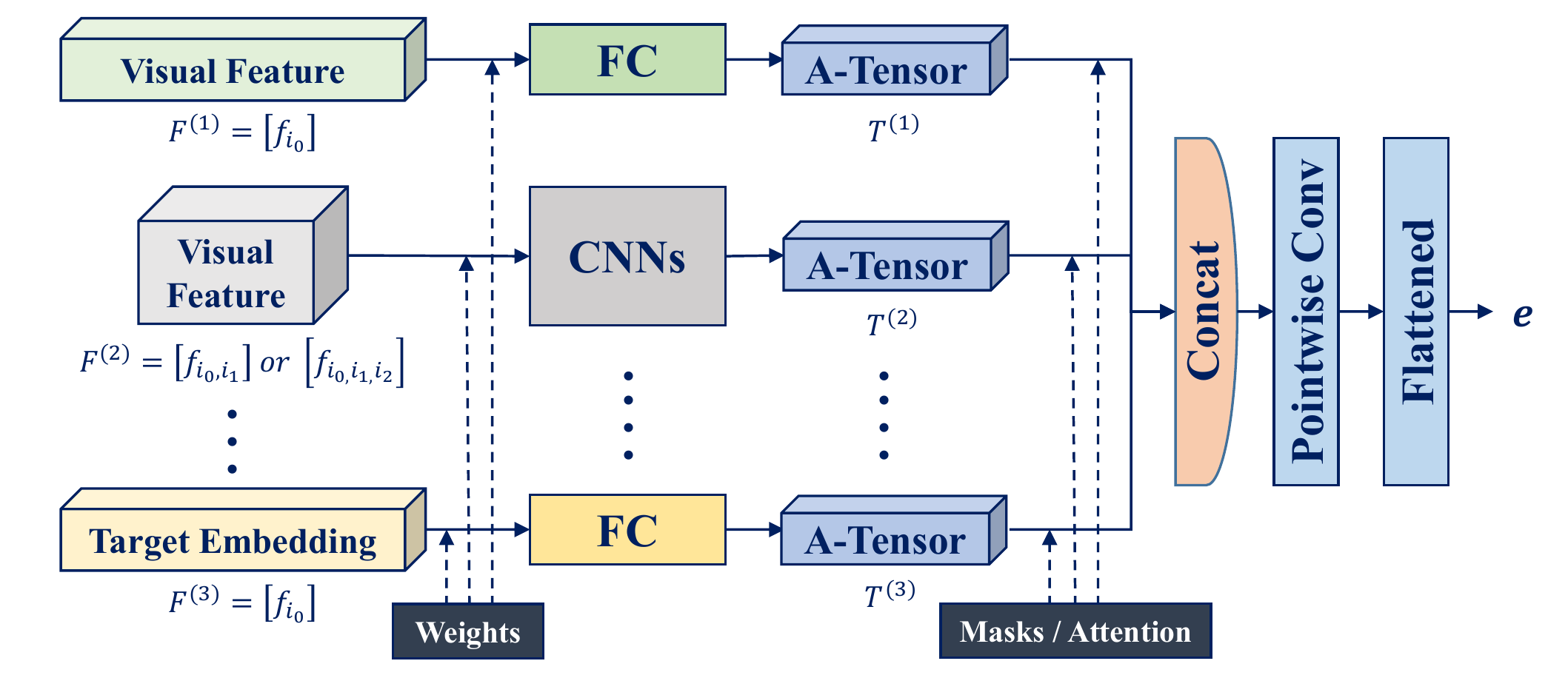}}
	\caption{Multi-processors and multimodal aggregation. 
 }
	\label{MultimodalFig}
\end{figure}

\indent Vision and hearing are two different modal perceptions. At the same time, information from a variety of different modalities can be analyzed from the visual observation. So how to deal with multi-modal data and aggregate them together is a key consideration for our model. \\
\indent Before aggregation, we need to convert these different features into A-Tensors, as shown in Fig.~\ref{MultimodalFig}. The details of the various visual feature encoders will be introduced in the experimental part. 

\indent Since the value distribution and shape of each feature are different, we adopt different modules as the multi-processors to handle different forms of weighted feature maps. Suppose we need to deal with a feature map $ F=[f_{i_{0},i_{1},...,i_{t}}]_{d_{0}\times d_{1}...\times d_{t}} \in \mathbb{R}^{d_{0}\times d_{1}...\times d_{t}} $ whose shape is a $ t $-dimensional vector $ (d_0,d_1,..,d_t) \in \mathbb{N}^t $. 

\indent In the case of $ t = 0 $, the feature map is a $ d_0 $-dimensional vector, then the linear layer is more suitable for the conversion module. Then the $ k $-channel aggregatable tensor converted from the vector is given by: 
\begin{equation}
    T=\mathrm{ReLU}(F \cdot W_{Linear} + B_{Linear})
\end{equation}{}where $ W_{Linear} \in \mathbb{R}^{d_{0} \times k}, B \in \mathbb{R}^{k} $ and the rectified linear unit (ReLU) is adopted as the activation function.

\indent When $ t = 2 $ or $ t =3  $, and if this feature map $ F $ has planar texture characteristics, shape characteristics and spatial relationship characteristics, we can use the convolution layers to further extract those characteristics which is helpful to the navigation system. The case of $ t = 2 $ can be regard as a special situation of $ t = 3 $ where $ d_0 = 1 $. We consider the $ d_0 $ dimension as the channel, then A-Tensor is obtained through one or more convolutional layers on the remaining two dimensions:
\begin{equation}
    T=\mathrm{ReLU}(G_{Conv}^{(\mu)} (F))
\end{equation}{}where $ G_{Conv}^{(\mu)} $ denotes $ \mu $-layer convolutional structure. In the initialization of these convolution layers, the weight is multiplied by a scaling factor due to the use of ReLU, which is recommended as $ \sqrt{2} $. 

\indent \textbf{Notation.} Denote the set of of natural numbers, $ n $-dimensional real vector space and $ n\times n $ dimensional real matrix space by $ \mathbb{N} $, $ \mathbb{R}^{n} $ and $ \mathbb{R}^{n\times n} $.

\indent To aggregate all those A-Tensors, we concatenate them and input them into a pointwise convolution layer, as shown in the Fig.~\ref{MultimodalFig}. We reshape the output of this convolution layer to a vector to obtain the final embedding $ e $ . 

\subsection{Region Attention Module}
\indent When people are walking or driving, they often see only local areas in the picture, such as obstacles, target objects, etc., which will affect the next action. We use the self-attention mechanism to select those areas in the observation which the agent really cares about during the navigation steps. The region attention dynamically attends on the region features to form a high-level impression of the visual content and feeds it to the memory network for decoding. 

\begin{figure}[htbp]
	\centerline{\includegraphics[width=3.1in]{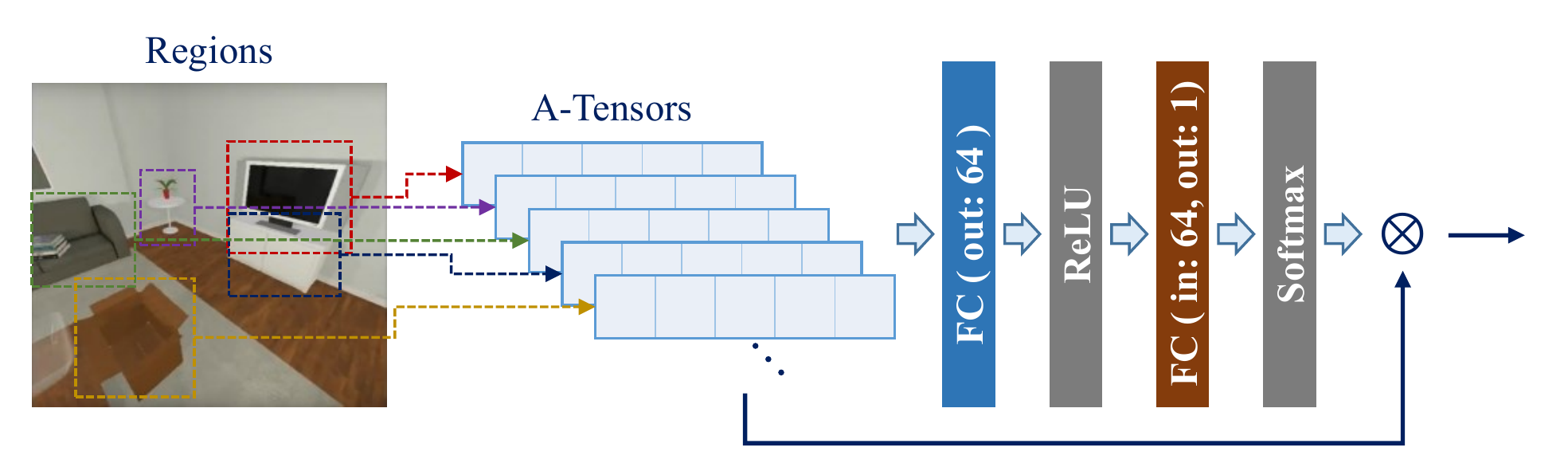}}
	\caption{The self-attention module for region features.}
	\label{AttentionFig}
\end{figure}

\indent The self-attention mechanism allows the inputs to interact with each other and find out who they should pay more attention to. In our region attention module, we use a double-layer linear layer to obtain the attention scores. As shown in Fig.~\ref{AttentionFig}, the softmaxed attention scores for each input is multiplied with its corresponding value.

\subsection{Learning Objective}
\indent The MVV-IN system optimize network parameters with two objective loss functions, one is called interaction loss $ \mathcal{L}_{int} $ and the another is navigation loss $ \mathcal{L}_{nav} $.  

\indent Our meta-learning approach is based on MAML \cite{finn2017model} algorithm, which optimize the network parameters on a huge set of tasks $ \mathcal{T}_{tr} $. Each task $ \tau_{tr} \in \mathcal{T}_{tr} $ has a tiny meta-training dataset and meta-validation dataset. In our navigation task, agent is designed to be continually learning as it interacts with the environment. So the meta-training dataset is defined as the observed images, hidden states and the issued actions for the first $ k $ steps of the agent's trajectory, denoted by $ \mathcal{D}_{\tau_{tr}}^{int} = (I_{<k}, h_{<k}, a_{<k}) $ according the section A, while the the meta-validation dataset is $ \mathcal{D}_{\tau_{tr}}^{nav} $, which represents the counterparts for the remainder of the trajectory. 

\indent During this interaction process, our algorithm modifies the navigation policy to minimize the interaction loss $ \mathcal{L}_{int} $, which is a self-supervised loss. In order to assist the agent to learn this objective, we structure a convectional neural network parameterized by $ \phi $ to compute it, which is denoted by $ \mathcal{L}_{int}^\phi $. This network is trained to imitate the $ \mathcal{L}_{nav} $, which is navigation loss defined with a reward for finding the object and a punishment for taking a step. Then the training objective for our MVV-IN system is given by:
\begin{equation}
    \mathop{\arg\max}_{\theta, \phi} \sum_{\tau_{tr} \in \mathcal{T}_{tr}} \mathcal{L}_{nav} (\theta-\psi \nabla_\theta \mathcal{L}_{int}^{\phi} (\theta, \mathcal{D}_{\tau_{tr}}^{int}), \mathcal{D}_{\tau_{tr}}^{nav})
\end{equation}{}where the network parameters $ \theta $ is optimized and updated to $ \theta-\psi \nabla_\theta \mathcal{L}_{int}^{\phi} (\theta, \mathcal{D}_{\tau_{tr}}^{int}) $ after $ k $ steps with respect to the self-supervised loss. 


\section{Experiments and Results}
\indent We articulate our experiments details and results in the following sections. 

\begin{figure*}[htbp]
	\centerline{\includegraphics[width=6.8in]{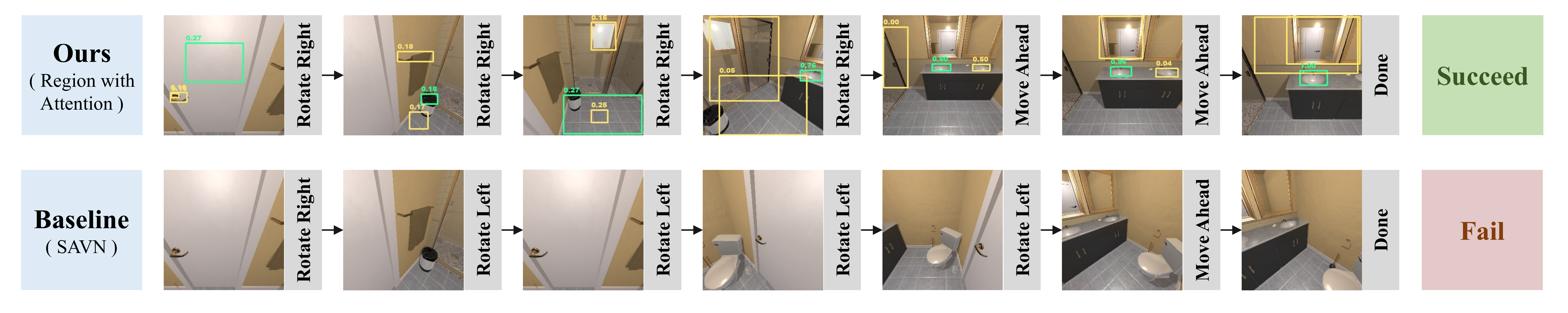}}
	\caption{Comparison with baseline in a sample navigation task: "Move to the sink".  }
	\label{CaseFig}
\end{figure*}

\subsection{Experiment Setup}

\indent \textbf{Dataset.} We evaluated the performance of our MVV-IN system and its variants on the SAVN dataset using the AI2-THOR environment \cite{kolve2017ai2}, which includes 4 different room types. Each room type provides 30 room instances. We use 20 rooms for training, 5 rooms for validation and testing respectively in each room type. We perform inference for 1000 difference tasks (50 for each room).

\indent \textbf{Auditory encoder.} To transfer voice to text, we use a novel open source speech recognition model called wav2letter++ \cite{pratap2018wav2letter++}. Next, to obtain the target from the command, we use the NLTK package (Natural Language Toolkit) \cite{Bird2009Natural} to recognize the noun word in the text.

\indent \textbf{Memory network.} Following the multimodal aggregation, the output is then flattened from the shape of $ (64, 7, 7) $ to a 64-dimentional vector, and given as input to a Long Short-Term Memory (LSTM) cell \cite{hochreiter1997long}. The hidden state size of LSTM is set to 512. 

\indent \textbf{Target objects.} We use the same target objects as SAVN\cite{wortsman2019learning}. In different room types, the target object set $ O $ is different. For example, in a kitchen, there are cooking utensils, such as a toaster and microwave, while in a bathroom, there are washing and body care supplies, like toilet paper and a liquid soap bottle. We embed the word of the target object with GloVe \cite{pennington2014glove}. 

\indent \textbf{Actions.} The action set $ A $ includes three movement-rotation commands, two camera commands and one "Done" signal. The movement-rotation commands are "Move Ahead", "Rotate Left" and "Rotate Right", and the camera commands are "Look Down" and "Look Up". The horizontal rotation occurs in steps of 45 degrees while the camera commands change the first-person camera incline angle by 30 degrees. Whether the agent completes a navigation task successfully or not, depends on whether, when the "Done' signal is raised, the target object is within 1 meter of the agent and within its field of view.

\indent \textbf{Training objective.} The navigation loss function $ \mathcal{L}_{nav}$ is defined with a reward of 5 for finding the object and a punishment -0.01 for taking a step. During the back propagation, Stochastic Gradient Descent (SGD) is used for the interaction gradient updates while Adam is adopted for navigation gradients. 

\subsection{Multimodal Feature Extraction and Conversion}

\indent Four different visual encoders are used to generate the corresponding visual features from the robot camera observation, as follows: 

\indent \textbf{RGB feature.} We use the ResNet-18 \cite{he2016deep} pretrained on ImageNet to extract a general image feature from the conv5\_4 layer. The shape of this feature is $ (512,7,7) $. 

\indent \textbf{Segmentation feature.} We employ Deeplab \cite{chen2017deeplab} to extract segmentation features. Specifically, the feature dimension of the last convolution layer is $ (2048,38,38) $. We average its second and third dimensions to get a 2048 dimensional vector. 

\indent \textbf{Depth feature.} MonoDepth \cite{godard2018digging} is adopted for depth features. Trained on Cityscapes \cite{cordts2016cityscapes} and KITTI \cite{geiger2013vision}, the depth feature map is extracted with a shape of $ (384, 512) $.\\ 
\indent \textbf{Region proposal and feature.} Faster-RCNN detector \cite{ren2015faster} with ResNeXt-101 backbone \cite{xie2017aggregated} is used for region proposal and feature extraction \cite{liu2020video}. The detector is pretrained on Visual Genome \cite{krishna2017visual}, and the region features are extracted from the fc6 layer. We select the 7 object regions with the highest confidence value, which are not background, each region feature is a 2048 dimensional vector. At the same time, we also consider to input the region proposal (position and size) into the network, which can be recorded as a 4-dimensional vector. 


\begin{table}[htbp]
\newcommand{\tabincell}[2]{\begin{tabular}{@{}#1@{}}#2\end{tabular}}
    \centering
    \begin{tabular}{|c|c|c|c|}
    \hline
         \textbf{Feature Type} & \textbf{\tabincell{c}{Dim \\ (In)}} & \textbf{Layer(s)} & \textbf{\tabincell{c}{Dim \\ (Out)}} \\ 
         \hline
         \textit{RGB} & 512 & $ 1 \times 1 $ conv, $ 1 \times 1 $ stride & 64  \\
         \hline
         \textit{Segmentation} & 2048 & FC & 64 \\
         \hline
         \textit{Depth} & 1 & \tabincell{c}{
             $ 3 \times 3 $ conv, $ 3 \times 3 $ stride; \\ 
             $ 3 \times 3 $ conv, $ 2 \times 3 $ stride; \\
             $ 3 \times 3 $ conv, $ 2 \times 3 $ stride; \\
             $ 3 \times 3 $ conv, $ 2 \times 1 $ stride; \\
             $ 3 \times 3 $ conv, $ 2 \times 2 $ stride \\
             } & 64 \\
         \hline
         \textit{Region (Feature)} & 2048 & FC; Self-Attention & 64 \\
         \hline
         \textit{Region  (Proposal)} & 4 & FC & 10 \\
         \hline
         \textit{Target} & 300 & FC & 64 \\
         \hline
         \textit{Action} & 6 & FC & 10 \\
         \hline
         
    \end{tabular}
    \caption{The architecture settings of multi-processors in our experiments. }
    \label{SettingsTable}
\end{table}{}
\indent To fuse the aforementioned features into A-Tensors, we applied different processors for each feature, whose architectures are illustrated in the Table~\ref{SettingsTable}. RGB, depth and segmentation features all use convolution layers in their conversion, while others use linear layers. For the A-Tensors of 7 regions, we use the self-attention mechanism to calculate the attention scores. 
\subsection{Evaluation Metrics}
\indent To evaluate our model, we use two metrics that illustrate the superiority of our method: 1. success rate;  and 2. success weighted by path length (SPL). Suppose that $ N $ is the number of episodes, $ S_i $, $ L_i $ and $ P_i $ denotes a binary indicator of success, the path length and the optimal trajectory length at episode $ i $ in that scene. Then, the success rate is defined as $ \frac{1}{N}\sum_{i=1}^N S_i $, while the SPL is defined as $ \frac{1}{N}\sum_{i=1}^N S_i\frac{L_i}{max(L_i,P_i)} $. 

\subsection{Quantitative Results}
\indent To compare the performance, we perform inference for 1000 different episodes (250 for each room type) for evaluation. Table~\ref{ResultsTable} illustrates the validation results of our various approaches versus the baseline method's results (SAVN). In all of these experiments, the specific feature is aggregated with the feature of general RGB as well as target word embedding. We provide results for all targets ‘All’ and a subset of targets whose optimal trajectory length $ L $ is greater than 5. 

\begin{table}[htbp]
\newcommand{\tabincell}[2]{\begin{tabular}{@{}#1@{}}#2\end{tabular}}
    \centering
    \begin{tabular}{|c|cc|cc|}
    \hline
         \multirow{2}*{} & \multicolumn{2}{|c|}{All} & \multicolumn{2}{|c|}{$ L\geq 5 $} \\
         \cline{2-5}
         & SPL & Success & SPL & Success \\
         \hline
         Baseline - SAVN & 13.86 & 40.20 & 12.27 & 25.25 \\
         \hline
         Segmentation & 15.52 & 45.50 & 12.46 & 27.24 \\
         \hline
         Depth & 14.85 & 43.40 & 12.69 & 26.41 \\
         \hline
         Region (Feature) & 15.26 & 46.00 & 12.03 & 28.07 \\
         \hline
         \tabincell{c}{Region (Feature) \\ with Self-Attention} & 16.04 & 46.50 & 12.02 & 27.74 \\
         \hline
         Region (Proposal) & 15.55 & 46.10 & \underline{13.83} & \underline{29.57} \\
         \hline
         \tabincell{c}{Segmentation \\ + Region (Proposal)} & \underline{17.04} & \underline{47.00} & \textbf{14.86} & \textbf{30.90} \\
         \hline
         \tabincell{c}{Segmentation \\ + Region (Feature)} & \textbf{17.27} & \textbf{48.70} & 13.63 & \textbf{30.90} \\
         \hline

    \end{tabular}
    \caption{Quantitative results (\%).  }
    \label{ResultsTable}
\end{table}{}

\indent From these metrics results, we can see that all of our methods outperform the baseline, which only uses RGB feature. Experiments show that using self-attention in region feature aggregation can improve the success rate and SPL for ‘All'. At the same time, the highest SPL for ‘$ L\geq 5 $' can be obtained by feeding the region proposal of object recognition into the network. In general, the best performance is provided by aggregating the features of RGB, segmentation, and region together. 

\subsection{Qualitative Demonstrations}
\indent A sample navigation task is shown in Fig.~\ref{CaseFig}. We have marked the top three regions of the self-attention score. The light green box is the one where the agent's attention is most focused on, as it has the highest self-attention score. It can be seen that without attention (using the baseline method), the agent loses direction and cannot recognize the target "sink" using the baseline method, while our method can quickly help the agent discover and locate the target and judge whether there are obstacles in front. 

\indent Note that except for the region corresponding to the target, the attention scores of other regions are fairly low. In addition, the scores of the two "sink" regions are both 0.5 when they first appear. As the agent approaches step by step, the score of the region corresponding to the nearest sink is higher. This illustrates how, by using region attention module, our MVV-IN system can effectively find the closest target object. More details are shown in the accompanying video: \url{https://youtu.be/sO_TIKu0TPk}.

\subsection{Time Efficiency and Abolition Study}
\begin{figure}[htbp]
	\centerline{\includegraphics[width=3.3in]{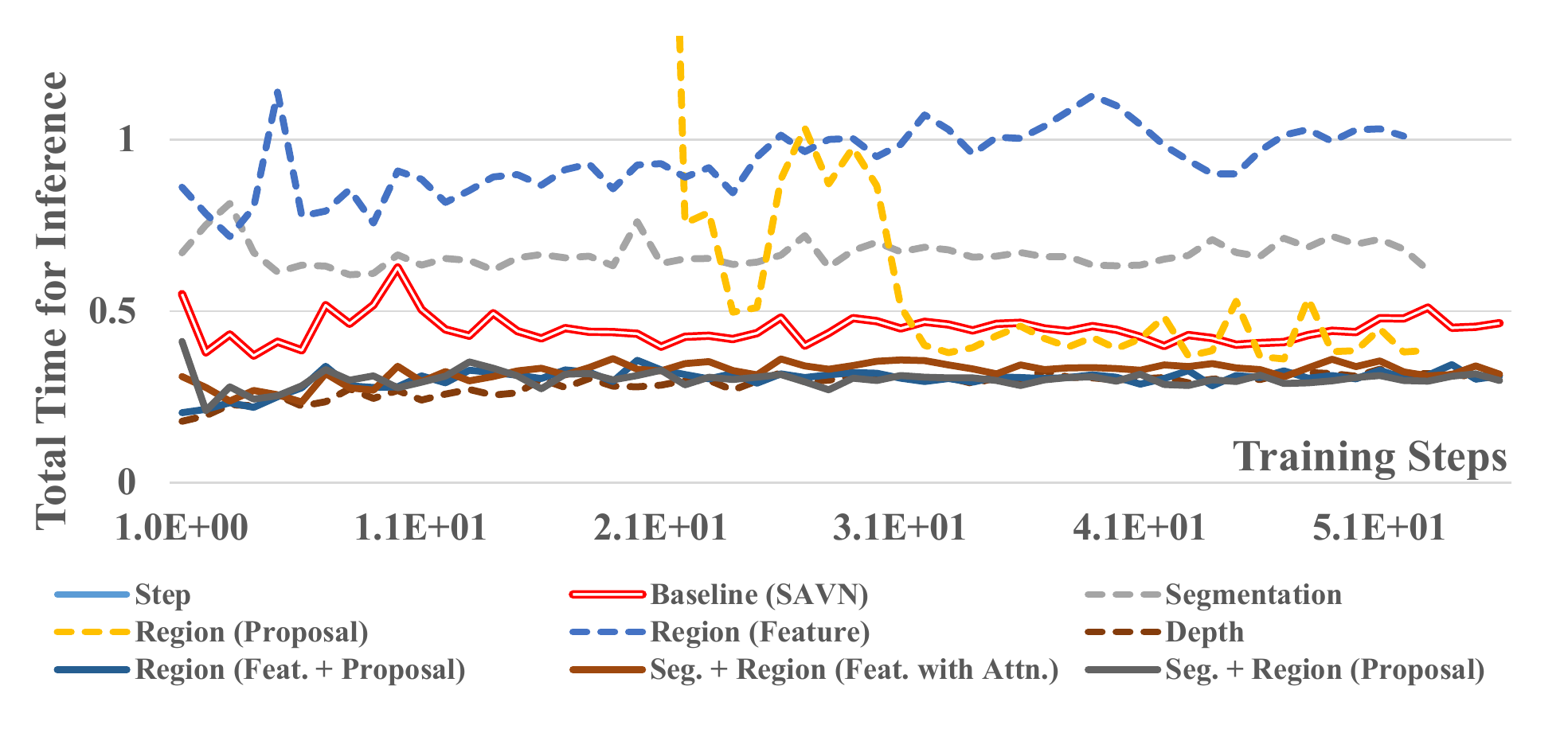}}
	\caption{Total inference time curve for evaluation. The solid line indicates that two features are aggregated into the network together.}
	\label{TimeFig}
\end{figure}

\indent As shown in Fig.~\ref{TimeFig}, aggregating the depth feature into the network, the region features together with the corresponding proposals, or the segmentation feature together with region proposals (or features with self-attention), all can reduce the inference time during the test. All the above experiments were performed on NVIDIA 2080 Ti GPUs. 

\indent Comparing the total time in Fig.~\ref{TimeFig}, we find that it is faster for inference to introduce the feature and proposal of the region into our network than if any of them would be introduced separately. Similarly, Table~\ref{ResultsTable} shows that adding both segmentation and region features achieves a better performance than adding one of them separately. All our models of variations outperform the baseline, which uses only RGB features. This illustrates the effectiveness of multimodal aggregation, including various visual information for indoor navigation. 
\section{Conclusion}
\indent In this study, we present a novel Memory Vision-Voice Indoor Navigation (MVV-IN) system. We use multimodal visual info to facilitate the environment's understanding in navigation systems, using only a monocular camera. Meta-learning technology is used during training so that our model is more robust to unseen scenes and novel tasks. We use extensive experiments to evaluate our method and demonstrate that it is competitive to the existing best approaches. Experimental results show that the proposed method achieves a higher success rate, SPL, and inference speed than the state-of-the-art baseline. In the future, we will evaluate the learning performance of 
our approach in real-world environments using a programmable mobile robot.  

\bibliographystyle{}

{\bibliography{test}}

\begin{thebibliography}{10}
\providecommand{\url}[1]{#1}
\csname url@samestyle\endcsname
\providecommand{\newblock}{\relax}
\providecommand{\bibinfo}[2]{#2}
\providecommand{\BIBentrySTDinterwordspacing}{\spaceskip=0pt\relax}
\providecommand{\BIBentryALTinterwordstretchfactor}{4}
\providecommand{\BIBentryALTinterwordspacing}{\spaceskip=\fontdimen2\font plus
\BIBentryALTinterwordstretchfactor\fontdimen3\font minus
  \fontdimen4\font\relax}
\providecommand{\BIBforeignlanguage}[2]{{%
\expandafter\ifx\csname l@#1\endcsname\relax
\typeout{** WARNING: IEEEtran.bst: No hyphenation pattern has been}%
\typeout{** loaded for the language `#1'. Using the pattern for}%
\typeout{** the default language instead.}%
\else
\language=\csname l@#1\endcsname
\fi
#2}}
\providecommand{\BIBdecl}{\relax}
\BIBdecl

\bibitem{anderson2018evaluation}
P.~Anderson, A.~Chang, D.~S. Chaplot, A.~Dosovitskiy, S.~Gupta, V.~Koltun,
  J.~Kosecka, J.~Malik, R.~Mottaghi, M.~Savva \emph{et~al.}, ``On evaluation of
  embodied navigation agents,'' \emph{arXiv preprint arXiv:1807.06757}, 2018.

\bibitem{liu2019virtual}
D.~Liu, Y.~Wang, K.~E. Ho, Z.~Chu, and E.~Matson, ``Virtual world bridges the
  real challenge: Automated data generation for autonomous driving,'' in
  \emph{2019 IEEE Intelligent Vehicles Symposium (IV)}.\hskip 1em plus 0.5em
  minus 0.4em\relax IEEE, 2019, pp. 159--164.

\bibitem{wortsman2019learning}
M.~Wortsman, K.~Ehsani, M.~Rastegari, A.~Farhadi, and R.~Mottaghi, ``Learning
  to learn how to learn: Self-adaptive visual navigation using meta-learning,''
  in \emph{Proceedings of the IEEE Conference on Computer Vision and Pattern
  Recognition}, 2019, pp. 6750--6759.

\bibitem{zhu2017target}
Y.~Zhu, R.~Mottaghi, E.~Kolve, J.~J. Lim, A.~Gupta, L.~Fei-Fei, and A.~Farhadi,
  ``Target-driven visual navigation in indoor scenes using deep reinforcement
  learning,'' in \emph{2017 IEEE international conference on robotics and
  automation (ICRA)}.\hskip 1em plus 0.5em minus 0.4em\relax IEEE, 2017, pp.
  3357--3364.

\bibitem{li2019walking}
J.~Li, S.~Tang, F.~Wu, and Y.~Zhuang, ``Walking with mind: Mental imagery
  enhanced embodied qa,'' in \emph{Proceedings of the 27th ACM International
  Conference on Multimedia}.\hskip 1em plus 0.5em minus 0.4em\relax ACM, 2019,
  pp. 1211--1219.

\bibitem{mousavian2019visual}
A.~Mousavian, A.~Toshev, M.~Fi{\v{s}}er, J.~Ko{\v{s}}eck{\'a}, A.~Wahid, and
  J.~Davidson, ``Visual representations for semantic target driven
  navigation,'' in \emph{2019 International Conference on Robotics and
  Automation (ICRA)}.\hskip 1em plus 0.5em minus 0.4em\relax IEEE, 2019, pp.
  8846--8852.

\bibitem{wang2018look}
X.~Wang, W.~Xiong, H.~Wang, and W.~Yang~Wang, ``Look before you leap: Bridging
  model-free and model-based reinforcement learning for planned-ahead
  vision-and-language navigation,'' in \emph{Proceedings of the European
  Conference on Computer Vision (ECCV)}, 2018, pp. 37--53.

\bibitem{wang2019reinforced}
X.~Wang, Q.~Huang, A.~Celikyilmaz, J.~Gao, D.~Shen, Y.-F. Wang, W.~Y. Wang, and
  L.~Zhang, ``Reinforced cross-modal matching and self-supervised imitation
  learning for vision-language navigation,'' in \emph{Proceedings of the IEEE
  Conference on Computer Vision and Pattern Recognition}, 2019, pp. 6629--6638.

\bibitem{savva2017minos}
M.~Savva, A.~X. Chang, A.~Dosovitskiy, T.~Funkhouser, and V.~Koltun, ``Minos:
  Multimodal indoor simulator for navigation in complex environments,''
  \emph{arXiv preprint arXiv:1712.03931}, 2017.

\bibitem{wu2018building}
Y.~Wu, Y.~Wu, G.~Gkioxari, and Y.~Tian, ``Building generalizable agents with a
  realistic and rich 3d environment,'' \emph{arXiv preprint arXiv:1801.02209},
  2018.

\bibitem{yu2018guided}
H.~Yu, X.~Lian, H.~Zhang, and W.~Xu, ``Guided feature transformation (gft): A
  neural language grounding module for embodied agents,'' \emph{arXiv preprint
  arXiv:1805.08329}, 2018.

\bibitem{gupta2017cognitive}
S.~Gupta, J.~Davidson, S.~Levine, R.~Sukthankar, and J.~Malik, ``Cognitive
  mapping and planning for visual navigation,'' in \emph{Proceedings of the
  IEEE Conference on Computer Vision and Pattern Recognition}, 2017, pp.
  2616--2625.

\bibitem{mirowski2016learning}
P.~Mirowski, R.~Pascanu, F.~Viola, H.~Soyer, A.~J. Ballard, A.~Banino,
  M.~Denil, R.~Goroshin, L.~Sifre, K.~Kavukcuoglu \emph{et~al.}, ``Learning to
  navigate in complex environments,'' \emph{arXiv preprint arXiv:1611.03673},
  2016.

\bibitem{nichol2018first}
A.~Nichol, J.~Achiam, and J.~Schulman, ``On first-order meta-learning
  algorithms,'' \emph{arXiv preprint arXiv:1803.02999}, 2018.

\bibitem{finn2017model}
C.~Finn, P.~Abbeel, and S.~Levine, ``Model-agnostic meta-learning for fast
  adaptation of deep networks,'' in \emph{Proceedings of the 34th International
  Conference on Machine Learning-Volume 70}.\hskip 1em plus 0.5em minus
  0.4em\relax JMLR. org, 2017, pp. 1126--1135.

\bibitem{torrey2010transfer}
L.~Torrey and J.~Shavlik, ``Transfer learning,'' in \emph{Handbook of research
  on machine learning applications and trends: algorithms, methods, and
  techniques}.\hskip 1em plus 0.5em minus 0.4em\relax IGI Global, 2010, pp.
  242--264.

\bibitem{weiss2016survey}
K.~Weiss, T.~M. Khoshgoftaar, and D.~Wang, ``A survey of transfer learning,''
  \emph{Journal of Big data}, vol.~3, no.~1, p.~9, 2016.

\bibitem{savinov2018semi}
N.~Savinov, A.~Dosovitskiy, and V.~Koltun, ``Semi-parametric topological memory
  for navigation,'' \emph{arXiv preprint arXiv:1803.00653}, 2018.

\bibitem{kahn2018self}
G.~Kahn, A.~Villaflor, B.~Ding, P.~Abbeel, and S.~Levine, ``Self-supervised
  deep reinforcement learning with generalized computation graphs for robot
  navigation,'' in \emph{2018 IEEE International Conference on Robotics and
  Automation (ICRA)}.\hskip 1em plus 0.5em minus 0.4em\relax IEEE, 2018, pp.
  1--8.

\bibitem{toshev2018visual}
A.~Toshev, A.~Mousavian, J.~Davidson, J.~Kosecka, and M.~Fiser, ``Visual
  representations for semantic target driven navigation,'' 2018.

\bibitem{wu2018learning}
Y.~Wu, Y.~Wu, A.~Tamar, S.~Russell, G.~Gkioxari, and Y.~Tian, ``Learning and
  planning with a semantic model,'' \emph{arXiv preprint arXiv:1809.10842},
  2018.

\bibitem{yang2018visual}
W.~Yang, X.~Wang, A.~Farhadi, A.~Gupta, and R.~Mottaghi, ``Visual semantic
  navigation using scene priors,'' \emph{arXiv preprint arXiv:1810.06543},
  2018.

\bibitem{liu2020video}
D.~Liu, Y.~Cui, Y.~Chen, J.~Zhang, and B.~Fan, ``Video object detection for
  autonomous driving: Motion-aid feature calibration,'' \emph{Neurocomputing},
  2020.

\bibitem{cummins2007probabilistic}
M.~Cummins and P.~Newman, ``Probabilistic appearance based navigation and loop
  closing,'' in \emph{Proceedings 2007 IEEE International Conference on
  Robotics and Automation}.\hskip 1em plus 0.5em minus 0.4em\relax IEEE, 2007,
  pp. 2042--2048.

\bibitem{wu2017scalable}
Y.~Wu, E.~Mansimov, R.~B. Grosse, S.~Liao, and J.~Ba, ``Scalable trust-region
  method for deep reinforcement learning using kronecker-factored
  approximation,'' in \emph{Advances in neural information processing systems},
  2017, pp. 5279--5288.

\bibitem{konolige2008outdoor}
K.~Konolige, M.~Agrawal, R.~C. Bolles, C.~Cowan, M.~Fischler, and B.~Gerkey,
  ``Outdoor mapping and navigation using stereo vision,'' in \emph{Experimental
  Robotics}.\hskip 1em plus 0.5em minus 0.4em\relax Springer, 2008, pp.
  179--190.

\bibitem{yu2018interactive}
H.~Yu, H.~Zhang, and W.~Xu, ``Interactive grounded language acquisition and
  generalization in a 2d world,'' \emph{arXiv preprint arXiv:1802.01433}, 2018.

\bibitem{zhang2016colorful}
R.~Zhang, P.~Isola, and A.~A. Efros, ``Colorful image colorization,'' in
  \emph{European conference on computer vision}.\hskip 1em plus 0.5em minus
  0.4em\relax Springer, 2016, pp. 649--666.

\bibitem{Sutskever2014Sequence}
I.~Sutskever, O.~Vinyals, and Q.~Le, ``Sequence to sequence learning with
  neural networks,'' \emph{Advances in Neural Information Processing Systems},
  vol.~4, 09 2014.

\bibitem{Cho2014On}
K.~Cho, B.~van Merriënboer, D.~Bahdanau, and Y.~Bengio, ``On the properties of
  neural machine translation: Encoder-decoder approaches,'' 09 2014.

\bibitem{kolve2017ai2}
E.~Kolve, R.~Mottaghi, D.~Gordon, Y.~Zhu, A.~Gupta, and A.~Farhadi, ``Ai2-thor:
  An interactive 3d environment for visual ai,'' \emph{arXiv preprint
  arXiv:1712.05474}, 2017.

\bibitem{pratap2018wav2letter++}
V.~Pratap, A.~Hannun, Q.~Xu, J.~Cai, J.~Kahn, G.~Synnaeve, V.~Liptchinsky, and
  R.~Collobert, ``wav2letter++: The fastest open-source speech recognition
  system,'' \emph{arXiv preprint arXiv:1812.07625}, 2018.

\bibitem{Bird2009Natural}
S.~Bird, E.~Klein, and E.~Loper, \emph{Natural Language Processing with
  Python}, 01 2009.

\bibitem{hochreiter1997long}
S.~Hochreiter and J.~Schmidhuber, ``Long short-term memory,'' \emph{Neural
  computation}, vol.~9, no.~8, pp. 1735--1780, 1997.

\bibitem{pennington2014glove}
J.~Pennington, R.~Socher, and C.~Manning, ``Glove: Global vectors for word
  representation,'' in \emph{Proceedings of the 2014 conference on empirical
  methods in natural language processing (EMNLP)}, 2014, pp. 1532--1543.

\bibitem{he2016deep}
K.~He, X.~Zhang, S.~Ren, and J.~Sun, ``Deep residual learning for image
  recognition,'' in \emph{Proceedings of the IEEE conference on computer vision
  and pattern recognition}, 2016, pp. 770--778.

\bibitem{chen2017deeplab}
L.-C. Chen, G.~Papandreou, I.~Kokkinos, K.~Murphy, and A.~L. Yuille, ``Deeplab:
  Semantic image segmentation with deep convolutional nets, atrous convolution,
  and fully connected crfs,'' \emph{IEEE transactions on pattern analysis and
  machine intelligence}, vol.~40, no.~4, pp. 834--848, 2017.

\bibitem{godard2018digging}
C.~Godard, O.~Mac~Aodha, M.~Firman, and G.~Brostow, ``Digging into
  self-supervised monocular depth estimation,'' \emph{arXiv preprint
  arXiv:1806.01260}, 2018.

\bibitem{cordts2016cityscapes}
M.~Cordts, M.~Omran, S.~Ramos, T.~Rehfeld, M.~Enzweiler, R.~Benenson,
  U.~Franke, S.~Roth, and B.~Schiele, ``The cityscapes dataset for semantic
  urban scene understanding,'' in \emph{Proceedings of the IEEE conference on
  computer vision and pattern recognition}, 2016, pp. 3213--3223.

\bibitem{geiger2013vision}
A.~Geiger, P.~Lenz, C.~Stiller, and R.~Urtasun, ``Vision meets robotics: The
  kitti dataset,'' \emph{The International Journal of Robotics Research},
  vol.~32, no.~11, pp. 1231--1237, 2013.

\bibitem{ren2015faster}
S.~Ren, K.~He, R.~Girshick, and J.~Sun, ``Faster r-cnn: Towards real-time
  object detection with region proposal networks,'' in \emph{Advances in neural
  information processing systems}, 2015, pp. 91--99.

\bibitem{xie2017aggregated}
S.~Xie, R.~Girshick, P.~Doll{\'a}r, Z.~Tu, and K.~He, ``Aggregated residual
  transformations for deep neural networks,'' in \emph{Proceedings of the IEEE
  conference on computer vision and pattern recognition}, 2017, pp. 1492--1500.

\bibitem{krishna2017visual}
R.~Krishna, Y.~Zhu, O.~Groth, J.~Johnson, K.~Hata, J.~Kravitz, S.~Chen,
  Y.~Kalantidis, L.-J. Li, D.~A. Shamma \emph{et~al.}, ``Visual genome:
  Connecting language and vision using crowdsourced dense image annotations,''
  \emph{International Journal of Computer Vision}, vol. 123, no.~1, pp. 32--73,
  2017.

\end{thebibliography}

\end{document}